\documentclass[runningheads]{llncs}

\usepackage{eccv}

\usepackage[dvipsnames]{xcolor}

\newcommand{\eblue}[1]{\color{eccvblue}#1}

\usepackage[utf8]{inputenc} %
\usepackage[T1]{fontenc}    %
\usepackage{url}            %
\usepackage{booktabs}       %
\usepackage{amsfonts}       %
\usepackage{nicefrac}       %
\usepackage{microtype}      %
\usepackage{xcolor}         %
\usepackage{graphicx}
\usepackage{multirow}
\usepackage{tabularx}
\usepackage{color}
\usepackage{colortbl}
\usepackage[export]{adjustbox}
\usepackage{placeins}	
\usepackage{enumitem}

\usepackage{pifont}
\newcommand{\cmark}{\ding{51}}
\newcommand{\xmark}{\ding{55}}

\newlength\tindent
\usepackage{eccvabbrv}

\usepackage{graphicx}
\usepackage{booktabs}

\usepackage[accsupp]{axessibility}  %

\usepackage[pagebackref,breaklinks,colorlinks,citecolor=eccvblue]{hyperref}

\usepackage{orcidlink}

\makeatletter
\renewcommand*{\@fnsymbol}[1]{\ensuremath{\ifcase#1\or *\or \dagger\or \ddagger\or
   \mathsection\or \mathparagraph\or \|\or **\or \dagger\dagger
   \or \ddagger\ddagger \else\@ctrerr\fi}}
\makeatother

\makeatletter
\newcommand{\printfnsymbol}[1]{%
  \textsuperscript{\@fnsymbol{#1}}%
}

\begin{document}

\title{Knowledge Distillation with Multi-granularity Mixture of Priors for Image Super-Resolution} 

\titlerunning{MiPKD}

\author{\hspace{-0.4cm}Simiao Li$^{1}$\thanks{Both authors contributed equally to this research} ~ Yun Zhang$^{1,2}$\printfnsymbol{1} ~ Wei Li$^{1}$  ~ Hanting Chen$^{1}$\\ 
    \hspace{-0.4cm}Wenjia Wang$^{2}$ ~ Bingyi Jing$^{3}$ ~ Shaohui Lin$^{4}$ ~ Jie Hu$^{1}$\thanks{Corresponding author: hujie23@huawei.com}\\
	$^{1}$Huawei Noah's Ark Lab\\
	$^{2}$Hong Kong University of Science and Technology (GZ)\\
	$^{3}$Southern University of Science and Technology\\
    $^{4}$East China Normal University\\
	{\tt\hspace{0cm}\{wei.lee, lisimiao\}@huawei.com}\quad\quad \tt{yzhangjy@connect.ust.hk}
}
\institute{}
\authorrunning{S. Li et al.}

\maketitle

\begin{abstract}  
Knowledge distillation (KD) is a promising yet challenging model compression technique that transfers rich learning representations from a well-performing but cumbersome teacher model to a compact student model. 
Previous methods for image super-resolution (SR) mostly compare the feature maps directly or after standardizing the dimensions with basic algebraic operations (e.g. average, dot-product). 
However, the intrinsic semantic differences among feature maps are overlooked, which are caused by the disparate expressive capacity between the networks. 
This work presents MiPKD, a multi-granularity mixture of prior KD framework, to facilitate efficient SR model  
through the feature mixture in a unified latent space and stochastic network block mixture. 
Extensive experiments demonstrate the effectiveness of the proposed MiPKD method.

\end{abstract}

\section{Introduction}
	
Super-resolution (SR) is a fundamental yet challenging task in the field of computer vision (CV), to restore high-resolution (HR) images from their low-resolution (LR) counterparts~\cite{dong2015image, liang2021swinir, chen2021pre}. In the past decade, convolutional neural network (CNN)~\cite{dong2014learning, kim2016accurate, lim2017enhanced, zhang2018image, dai2019second} and Transformer~\cite{chen2021pre, liang2021swinir,wang2022uformer,zamir2022restormer,zhang2024distilling} have demonstrated exceptional success for SR. 
However, it is impractical to directly deploy these models on resource-limited devices due to their heavy computation overload~\cite{zhang2021aligned}. 
Consequently, there has been a growing interest in model compression for SR models to facilitate their real-world applications.

 \begin{figure*}[!t]
		\centering
		\includegraphics[width=1\textwidth,height=0.28\textwidth]{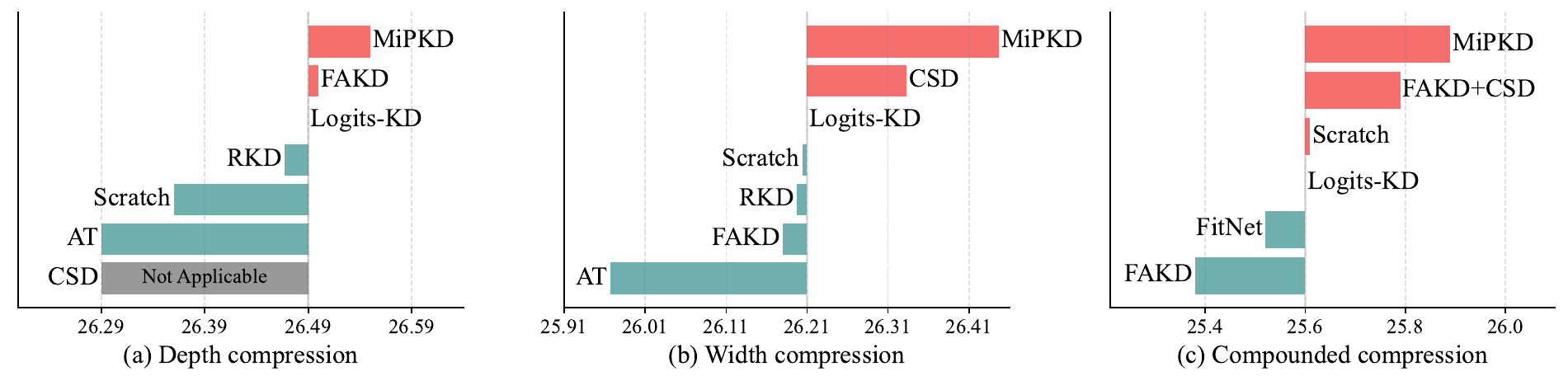}
		\vspace{-3mm}
		\caption{The PSNR of student models on Urban100 testset under different compression settings. In the depth compression (a), there are barely KD methods outperforming vanilla logits-KD. For width compression (b), CSD performs well but only satisfies this setting. For compounded compression, almost all KD underperforms training without KD.
  }
		\label{fig:motivation}
		\vspace{-5mm}
	\end{figure*}

Knowledge distillation, emerging as an effective model compression method, can significantly reduce the computation overload, facilitating the student by transferring dark knowledge from the well-performed but cumbersome teacher model to the compact one~\cite{zhang2021data, luo2021boosting, hui2019lightweight, lee2020learning}. Compared with other model compression techniques, such as quantization~\cite{li2020pams, hong2022cadyq, ma2019efficient},  pruning~\cite{wang2021exploring,wang2021towards}, compact block design~\cite{ahn2018fast,song2021addersr,nie2021ghostsr,wang2022efficient}, and neural architecture search~(NAS)~\cite{zoph2016neural, wan2020fbnetv2, ren2021comprehensive},  KD is a widely recognized method that can be combined with aforementioned techniques to further improve the compactness of student model.
KD for SR has also attracted wide attention recently and gained remarkable progress~\cite{li2020pams, lee2020learning, zhang2021data, he2020fakd, wang2021towards, zhang2023data}. These methods can be roughly categorized into response-based KD and feature-based KD, the former utilizes the output of the teacher model to supervise the training process of the student model, while the latter tries to align the 
hidden representations between the teacher model and the student model~\cite{gou2021knowledge, wang2021towards, he2020fakd}.

Though previous KD methods show promising results in SR, several issues hinder the wide applications of them. First, existing KD techniques for SR are tailored to specific teacher-student architectures. They support either network depth (\cref{fig:motivation} (a)) or network width~(\cref{fig:motivation} (b)) compression~\cite{he2020fakd}, and deteriorates the student dramatically when adopting them into another setting. For instance, FAKD~\cite{he2020fakd} boosts the student model in depth compression but deteriorates the student when applied to a width compression circumstance. CSD~\cite{wang2021towards} improves the student model significantly~(Fig.\ref{fig:motivation} (b)) but is not compatible with depth compression in Fig.~\ref{fig:motivation} (a). 
It's necessary to propose a more flexible KD framework which is closer to real-world application. 
While few methods have discussed the compounded compression on both depth and width dimension, which is a much more general but challenging scenario. 
To be specific, FAKD and those feature-based KD methods introduced from high-level CV, \textit{e.g.} RKD~\cite{park2019relational}, AT~\cite{zagoruyko2016paying}, and FitNet~\cite{Romero2014FitNetsHF} hardly benefit the student model.
\cref{fig:motivation} (a) shows that the previous depth and channel distillation methods can just obtain a marginal performance gain or even deteriorate the student in most cases. 
To alleivate these issues, in this paper, we present a novel knowledge distillation framework for SR models, the multi-granularity mixture of prior knowledge distillation (MiPKD), that is universally applicable to a wide array of teacher-student architectures at feature and block levels.
    Specifically, the feature prior mixer first dynamically combines priors from the teacher and the student and then is 
    supervised by the teacher’s feature map. The block prior mixer adopts a coarser-grained prior mixture at the network block level that dynamically switches the normal forward propagation path to the teacher or the student. The final output of this newly ensembled sub-network is further supervised by the teacher's response. 
    
    In summary, the main contributions of this paper are as follows: 
	\begin{itemize} 
		\item[$\bullet$] We present MiPKD, a KD framework for efficient SR, transferring the teacher model's prior knowledge from both network width and depth levels. It's more flexible and applicable to a wide array of teacher-student architectures.

        \item [$\bullet$] We propose the feature and block prior mixers to reduce the capacity disparity between teacher and student models for better alignment. The former combines the feature maps in a unified latent space, while the latter assembles dynamic combination of network blocks from teacher and student models. 
		\item[$\bullet$] Extensive experiments on various benchmarks show that the proposed MiPKD framework significantly outperforms the previous arts.
	\end{itemize}

\section{Related Work}
	\textbf{Deep SISR Models.} Deep neural networks~(DNNs) based image super-resolution have shown impressive success. Dong \textit{et al.}~\cite{dong2014learning} firstly introduced CNN with only three Conv layers for image SR. Residual learning was introduced in VDSR~\cite{kim2016accurate}, reaching 20 Conv layers. Lim \textit{et al.} proposed EDSR~\cite{lim2017enhanced} with simplified residual block~\cite{ResNet}. Zhang \textit{et al.} proposed an even deeper network RCAN~\cite{zhang2018image}. 
	Later, Mei \textit{et al.} proposed CSNLN~\cite{zhang2019residual} by combining feature correlations, and external statistics. Most of them have achieved state-of-the-art results with deeper and wider networks. Recently, Transformer has gained a surge of interest for image restoration. Chen \textit{et al.} proposed the first pre-trained image processing transformer IPT~\cite{chen2021pre}. 
	SwinIR~\cite{liang2021swinir} first applies the residual Swin Transformer block to the image restoration for deep feature extraction.
	Restormer~\cite{zamir2022restormer} proposed a multi-scale hierarchical design that incorporates efficient Transformer blocks by modifying self-attention and MLP. 
	Similarly, Uformer~\cite{wang2022uformer} proposed a LeWin Transformer block for image restoration. While CNNs and Transformers have demonstrated impressive performance in SISR, they suffer from heavy memory cost and computational overhead. \newline
    \textbf{Efficient SISR.}
	To improve the model efficiency, various approaches have been proposed to reduce model's redundancy, such as  neural architecture search (NAS)~\cite{chu2021fast, song2020efficient}, compact block design~\cite{ahn2018fast,song2021addersr,nie2021ghostsr,wang2022efficient,wang2022uformer,zamir2022restormer}, pruning~\cite{wang2021exploring,wang2021towards}, and low-bit quantization~\cite{ma2019efficient,li2020pams,hong2022cadyq}. The strength of NAS manifests in searching the optimal architecture but is time-consuming and computationally expensive due to the massive search space. 
	Afterwards, compact SR model designs have attracted rising attention and achieved remarkable progress~\cite{zhang2022efficient, hui2019lightweight, ahn2018fast, dong2016accelerating}.
	Zhang \textit{et al.} proposed ELAN~\cite{zhang2022efficient} that designs a group-wise multi-scale self-attention~(GMSA) module to exploit the long-range image dependency, achieving even better results against the transformer-based SR models but with significantly less complexity.
	Pruning~\cite{wang2021exploring,wang2021towards} and quantization~\cite{ma2019efficient,li2020pams,hong2022cadyq} are other two types of methods to remove model redundancy by sparsity and low-bit quantization mappings. Although those lightweight networks have achieved great progress, they still need considerable extra computation resources.\newline 
    \noindent \textbf{Knowledge Distillation for SISR.}\label{sec: related-sisr-kd}
	Knowledge distillation is widely recognized as an effective model compression method that can significantly reduce the computation overload and improve student's capability by transferring dark knowledge from the large teacher model to the lightweight student model~\cite{gou2021knowledge, yim2017gift, hinton2015distilling}. 
	Recently, several attempts have also been made for image super-resolution knowledge distillation. 
	Lee \emph{et al}.~\cite{lee2020learning} first pre-trained the encoder and decoder using the pairs of the same HR images, and the privileged information is extracted from the decoder to obtain the statistical output location and scale maps as the knowledge.
	He \emph{et al.} proposed FAKD~\cite{he2020fakd} to distill the second-order statistical information from the affinity matrix of feature maps. 
	Wang \emph{et al.} proposed CSD that incorporates self-distillation and contrastive learning~\cite{wang2021towards} by introducing extra simply upsampled LR images as negative samples.
	However, none of the existing SRKD methods have discussed how to customize a proper teacher for a student with limited capacity or whether a stronger teacher consistently benefits the capacity-limited student.
	Furthermore, existing SRKD techniques for SR are tailored to specific teacher-student architectures, focusing on either network depth~\cite{wang2021towards} or channel compression~\cite{he2020fakd}, which is infeasible for practical compounded compression applications. %

\section{Methodology}
\label{headings}

\begin{figure*}[t]
    \centering    \includegraphics[width=\linewidth]{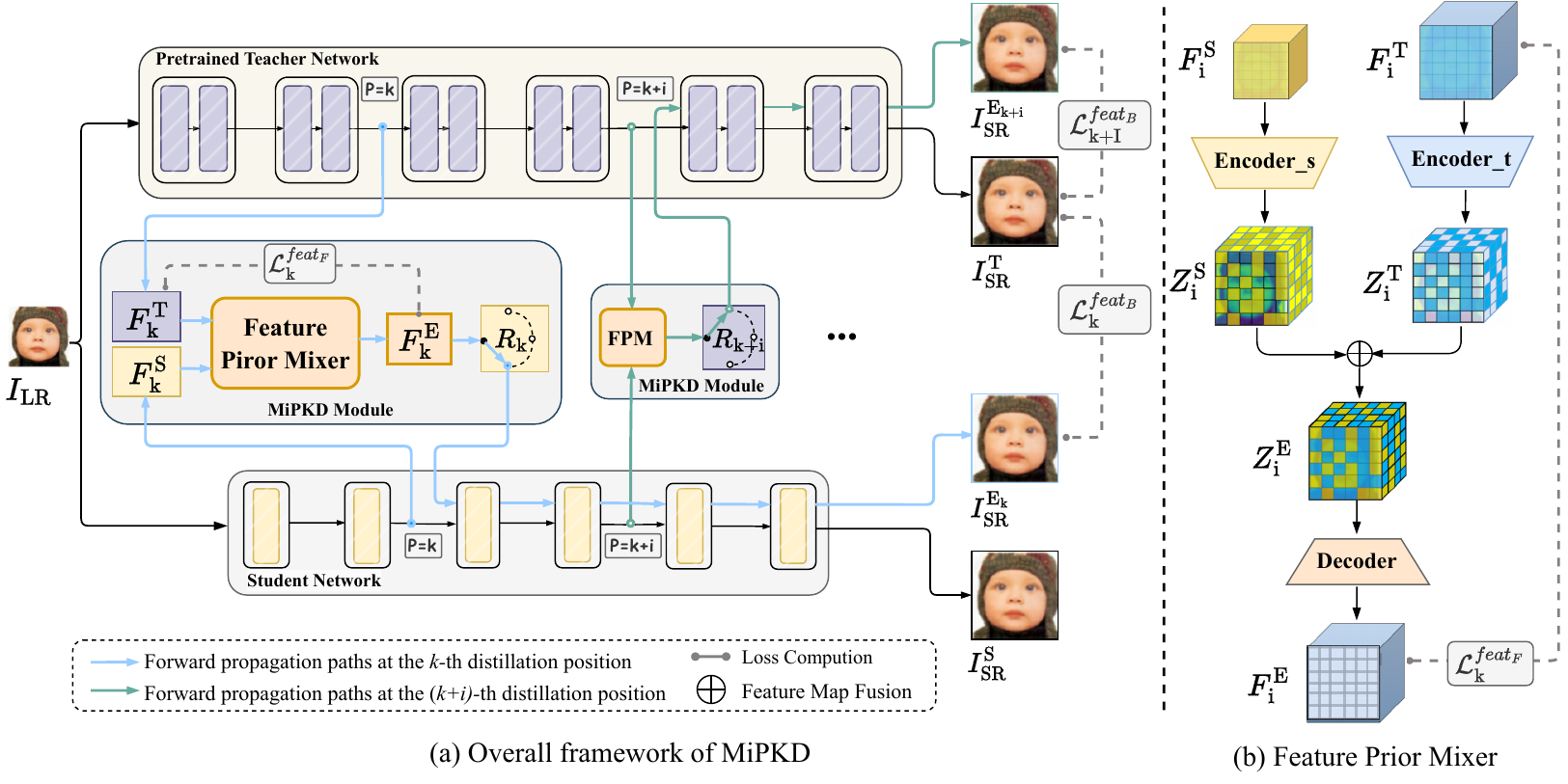}\vspace{-0.75em}
    \caption{Framework of the MiPKD method. %
    MiPKD utilizes the multi-granularity prior mixture to constrain the KD process. 
    The feature prior mixer dynamically combines priors from the teacher and student, and the block prior mixer adopts a coarser-grained prior mixture at the network block level. 
    }\vspace{-1em}
    \label{fig:mipkd}
\end{figure*}
 
\subsection{Preliminaries and Notations} 
Given a low-resolution input image $I_{LR}$, the deep SR model $\mathcal{F}(\cdot)$ aims to reconstruct the high-resolution image $I_{SR} = \mathcal{F}(I_{LR};\Theta)$ with fine details and looking similar with corresponding high-resolution image $I_{HR}$, where $\Theta$ is the model parameters. 
Existing KD methods for SR can be divided into two categories: logits-based and feature-based. 
The former compels the student model $\mathcal{F}_{S}$ to produce the same prediction as the teacher model
\begin{equation}
	\mathcal{L}_{logits} = \mathcal{D}_{logits}(I_{SR}^S, ~I_{SR}^T)
	\label{eq1}
\end{equation}
where $I_{SR}^S=\mathcal{F}_{S}(I_{LR};\Theta^{S})$ and $I_{SR}^T=\mathcal{F}_{T}(I_{LR};\Theta^{T})$ represent the output SR images of the student and teacher models, and $\mathcal{D}_{logits}$ is the loss function that measures the difference between two models' outputs, \textit{e.g.} the $L_1$ or $L_2$ loss function. 
Similarly, the feature-based KD methods aim to mimic the rich implicit learning representations between the teacher and the student, which also can be represented as an auxiliary loss item:
\begin{equation}
	\mathcal{L}_{feat} = \mathcal{D}_{feat}(\mathcal{T}_s(\mathrm{F}^S_k), ~\mathcal{T}_t(\mathrm{F}^T_k))
	\label{eq2}
\end{equation}
where $\mathrm{F}^S_k$ and ${\mathrm{F}}^{T}_k$ denote the feature maps of the student model and the teacher model at the $k$-th distillation position, respectively. $\mathcal{T}_t$ and $\mathcal{T}_s$ 
are the transformations applied on raw feature maps and $\mathcal{D}_{feat}$ is the loss function for feature distillation.

\subsection{Mixture of Prior Knowledge Distillation}

\noindent According to earlier research and analysis, there is a negative correlation between student performance and the teacher's model's capacity as the capacity gap widens between students and teachers~\cite{mirzadeh2020improved}. Since the student models struggle to grasp the capacity to extract semantic information from the bigger model. In the process of feature distillation, it is crucial to illustrate the effective retrieval and transfer of the teacher's prior knowledge to the student network. 
Our proposed feature distillation method, 
surpasses FAKD, CSD and other deep or channel kd methods as Fig. \ref{fig:motivation} shown. This is accomplished by incorporating the teacher's knowledge with the student's network at feature and block levels, within a stochastic dynamic framework. This approach reduces the capacity disparity between the teacher and student, thereby enhancing the performance of the student model.

\noindent \textbf{Feature Prior Mixer.}  

\noindent Fig. \ref{fig:mipkd} illustrates the hybrid prior knowledge framework at the feature level, including the encoder modules for the teacher and student feature maps and the decoder module for the fused feature map. At the $k$-th feature distillation position, initially, the feature maps of both the student model 
$\mathrm{\textbf{F}}^S_k$ 
and teacher model $\mathrm{\textbf{F}}^T_k$ are fed into the respective encoder models to obtain the representations
$\mathrm{\textbf{Z}}^S_k$, $\mathrm{\textbf{Z}}^T_k\in{\mathbb{R}}^{C\times H\times W}$in a unified latent space, where $C, H, W$ are the dimension of the feature maps. 
Then the encoded feature maps are fused according to a random 3D-mask $\mathbf{I}^M \in \{0,1\}^{C\times H\times W}$ as
\begin{equation}
F^E_k = D(Z^S_k\odot(1-\mathbf{I}^M) + \mathbf{Z}^T_k\odot(\mathbf{I}^M)),
	\label{mask_stitch}
\end{equation}
where $D$ is a decoder that reverts the fused feature map as the enhanced feature map representation $\mathrm{\textbf{F}}^E_k$, and $\odot$ denotes the element-wise product between matrices.
The student's feature map is combined with the teacher's prior knowledge with an encoder-decoder structure to reduce the discrepancy between them at the feature level. $\mathbf{F}^E_k$ is subsequently utilized as an input to the subsequent block level prior mixer module.
The feature distillation loss $\mathcal{L}^{feat_F}_{k}$ of Feature Prior Mixer is computed between $\mathbf{F}^E_k$ and $\mathbf{F}^T_k$ as
 \begin{equation}
	\mathcal{L}^{feat_F}_k = \mathcal{D}_{feat}(\mathrm{\textbf{F}}^{E}_k, ~\mathcal\mathrm{\textbf{F}}^T_k)
	\label{loss_f}
\end{equation}
Additionally, in order to enhance the reconstruction capability of the decoder and ensure the stability of the training process,  
the auxiliary enhanced feature map $\mathrm{\textbf{F}}^{'E}_k$ is obtained by directly inputting the teacher's feature map to the teacher's encoder and decoder without applying masking and mixing strategy. The auxiliary ``auto-encoder'' loss $\mathcal{L}^{ae}_k$ is computed
as
 \begin{equation}
	\mathcal{L}^{ae}_k = \mathcal{D}_{feat}(\mathrm{\textbf{F}}^{'E}_k, ~\mathcal\mathrm{\textbf{F}}^T_k).
	\label{loss_f2}
\end{equation}
It requires the encoder and decoder to serve as an auto-encoder structure, ensuring the decoded enhanced feature map is comparable with $\mathbf{F_k^T}$. The enhancement of the decoder contributes to the overall effectiveness of the feature prior mixer module.

\noindent \textbf{Block Prior Mixer.}

\noindent Existing feature-based distillation methods on SR tasks mostly align the feature representations in the original space with Mean Absolute Error (MAE) or Mean Square Error (MSE). 
Previous studies~\cite{pasqualetti2014controllability} shows that distinct controllability properties and semantics may exist among the network of the similar architecture with various weights. 
Student and teacher models with various dimensional space typically exist seperate distribution of semantic information~\cite{liu2023function}. 
Therefore, solely aligning features at the present distillation node with the same magnitudes of distance can leads the student model to learn entirely different information. 
To tackle this issue, we propose to align the networks' ability of processing and representing information by assembling a dynamic combination of blocks and transmitting the fusion information from the Feature Prior Mixer to the enhanced network.

To construct an enhanced network ($\mathcal{F}^{block}_{E}$) at distillation position $k$, according to the Block Prior Mixing Option $\mathrm{R}_k$ and  $\mathrm{D}_k$ randomly sampled from $\{0,1\}$, the output of Feature Prior Mixer $\mathbf{F}^E_k$ is forwarding propagated to the student network ($\mathrm{R}_k=1$,$\mathrm{D}_k=1$), teacher network ($\mathrm{R}_k=0$,$\mathrm{D}_k=1$), or deposed ($\mathrm{D}_k=0$), as the dashed path exemplified in Figure \ref{fig:mipkd}.  $\mathcal{B}_{{S}_{(k)}}$ and $\mathcal{B}_{{T}_{(k)}}$ represent the block from student and teacher models at the current position respectively.
$\mathcal{B}_{{O}_{(k)}}$ represents the mixed block at the current position based on $\mathrm{R}_k$ and $\mathrm{D}_k$ as formula \ref{bo} shown.
\begin{equation}
	\mathcal{B}_{{O}_{(k)}}=\mathrm{D}_k(\mathrm{R}_k\mathcal{B}_{{S}_{(k)}} + (1-\mathrm{R}_k)\mathcal{B}_{{T}_{(k)}}).
	\label{bo}
\end{equation}
Based on this process, denote the output of such concatenated network as $I^{E}_{SR}$,
\begin{equation}
	{I}_{SR}^E=\mathcal{F}^{block}_{E}(I_{LR};\Theta^{S})={\mathcal{B}_{{O}_{(k)}}}(\mathbf{F}^E_k)
	\label{fb}
\end{equation}
The feature knowledge distillation loss based on Block Prior Mixer is derived through the ultimate combined network output with the teacher output and ground truth as formula \ref{loss_b3} shown, where $w$ represents the trade-off between the supervisions from teacher model and ground-truth (GT). For the SR tasks, w equals to 0.5 in the following experiments.
 \begin{equation}
    {L}^{feat_B}_k  = w \mathcal{D}(I_{SR}^E, ~I_{SR}^T) + (1-w)\mathcal{D}(I_{SR}^E, ~I_{HR})
	\label{loss_b3}
\end{equation}
In addition, $L_k^{feat_B} $ is ignored if the $k$-th feature distillation position is dropped out. It is anticipated that there will be an attainment of interchangeability between the corresponding teacher and student network blocks, allowing the student to inherit and replicate the capabilities of the teacher model.

\noindent\textbf{The Whole Pipeline.}  

\noindent In general, 
for each feature distillation position $k$, based on the pair of $\mathbf{F}^T_k$ and $\mathbf{F}^S_k$ as the input of Feature Prior Mixer, the random masked feature map of student is populated and fused with the teacher's as ${\mathbf{F}}^E_k$, and the ${L}^{feat_E}_{k}$ is computed to align with the initial teacher feature map in the same distribution space.
Meanwhile, $\mathrm{R}_{k}$ and $\mathrm{D}_{k}$ determine the transmission direction of ${\mathbf{F}}^E_k$  for each iteration, the students' network for random block information is also exchanged and transmitted by the teacher's as Fig.\ref{fig:mipkd} shown. 
Besides logits-KD loss ${L}_{logits}$, reconstruction loss ${L}_{rec}$, the feature losses in block level and feature level are accumulated:
 \begin{equation}
    {L}_{total}  = {\lambda_{kd}}{L}_{logits} +   {\lambda_{rec}}{L}_{rec} +\sum_{k\leq{K}}({\lambda_{feat}}{L}^{feat_E}_{k} +{\lambda_{block}}{L}^{feat_B}_{k}).
	\label{loss_b3}
\end{equation}
where $\lambda_{rec}$, $\lambda_{kd}$, $\lambda_{feat}$, $\lambda_{block}$  represent the weights for reconstruction loss, logits-kd loss, feature prior mixer and block prior mixer respectively.
The teacher's prior knowledge is effectively transferred through this multi-level distillation process. The following experiments illustrate the efficiency of the above strategy.

\begin{table}[t]	\caption{SR model specifications on $\times$4 experimental settings. The {\#}Params, FLOPs and FPS are calculated with a 256$\times$256$\times$3 input image and FPS is computed on a single NVIDIA V100 GPU. %
		}\vspace{-0.5em}	
		\label{tab:model-config}	\centering	\small 
		\resizebox{\columnwidth}{!}{	
            \setlength{\tabcolsep}{0.2em}
		\begin{tabular}{@{\hspace{2pt}}lccccccc@{\hspace{2pt}}}
			\toprule
			\multirow{2}{*}{Model}
			& \multirow{2}{*}{Role} & \multicolumn{3}{c}{Network}
			& \multirow{2}{*}{FLOPs (G)} & \multirow{2}{*}{\#Params (M)}  & \multirow{2}{*}{FPS} \\
			\cmidrule(lr){3-5} &  & Channel & Block & Group &  &  &  \\ 
			\midrule		
			\multirow{3}{*}{EDSR}  & Teacher  & 256  & 32  & -  & 3293.35  & 43.09   & 3.2 \\
			& Student 1  & 64  & 32    & -   & 207.28 &  2.70  & 33.958 \\
			& Student 2 & 64  & 16  & -  & 129.97 (25.3$\times$) & 1.52 (28.3$\times$) & 53.3 \\
			\midrule
			\multirow{2}{*}{RCAN}  & Teacher  & 64 &  20  & 10 & 1044.03  & 15.59 & 6.3 \\
			& Student & 64 & 6  & 10  & 366.98 & 5.17 & 12.3 \\ 
			\midrule
			\multirow{2}{*}{SwinIR} & Teacher               & 180        &   6       &    -      & 861.27                    & 11.90 M                   &  0.459               \\		& Student               & 60        &      4     &   -       & 121.48                     & 1.24 M                    &   0.874             \\
			\bottomrule	
	\end{tabular} }
\end{table}

\section{Experimental Results}
	\subsection{Experiment Setups}
\textbf{Backbones and Evaluation.} We use EDSR~\cite{lim2017enhanced}, RCAN~\cite{zhang2018image}, and SwinIR~\cite{liang2021swinir} as backbone models to verify the effectiveness of MiPKD and compare it with prior KD methods on $\times2$, $\times3$, and $\times4$ super-resolving scales. 
The SR network specifications and some statistics are presented in~\cref{tab:model-config}, including the number of channels, residual blocks and residual groups (RCAN), number of parameters~({\#}Params), FLOPs, and inference speed~(frame per second, FPS). To evaluate the KD methods under high compression rate, we trained an extra EDSR student network.

\begin{table}[htb]
\caption{Quantitative comparison of distilling EDSR~\cite{lim2017enhanced} on the benchmark datasets. In these experiments, the EDSR student model of c64b32 is distilled by the teacher model of c256b32.}
\label{tab:mipkd-edsr}
\centering
	\begin{tabular}{@{}l@{\hspace{3\tabcolsep}}l@{\hspace{3\tabcolsep}}c@{\hspace{3\tabcolsep}}c@{\hspace{3\tabcolsep}}c@{\hspace{3\tabcolsep}}c@{}}
		\toprule
		\multirow{2}{*}{Scale} & \multirow{2}{*}{Method} & Set5         & Set14        & BSD100       & Urban100     \\\cmidrule(l){3-6} 
		&                         & PSNR/SSIM    & PSNR/SSIM    & PSNR/SSIM    & PSNR/SSIM    \\ \midrule
		\multirow{9}{*}{x2}    & Teacher                 & 38.20/0.9606  & 34.02/0.9204 & 32.37/0.9018 & 33.10/0.9363  \\
		& Scratch           & 38.00/0.9605    & 33.57/0.9171 & 32.17/0.8996 & 31.96/0.9268 \\
		& KD         & 38.04/0.9606 & 33.58/0.9172 & 32.19/0.8998 & 31.98/0.9269 \\
		& RKD                     & 38.03/0.9606 & 33.57/0.9173 & 32.18/0.8998 & 31.96/0.9270  \\
		& AT                      & 37.96/0.9603 & 33.48/0.9167 & 32.12/0.8990  & 31.71/0.9241 \\
		& FitNet                  & 37.59/0.9589 & 33.09/0.9136 & 31.79/0.8953 & 30.46/0.9111 \\
		& FAKD             & 37.99/0.9606 &
  33.60/0.9173 &
  32.19/0.8998 &
  32.04/0.9275  \\
		& CSD              & 38.06/0.9607 &
  33.65/0.9179 &
  32.22/0.9004 &
  32.26/0.9300 \\
		& \eblue{MipKD}                   & \eblue{38.16/0.9611}              &     \eblue{33.85/0.9194}         &   \eblue{32.27/0.9008}           &      \eblue{32.52/0.9318}        \\ \midrule
		\multirow{9}{*}{x3}    & Teacher                 & 34.76/0.929  & 30.66/0.8481 & 29.32/0.8104 & 29.02/0.8685 \\
		& Scratch           & 34.39/0.927  & 30.32/0.8417 & 29.08/0.8046 & 27.99/0.8489 \\
		& KD         & 34.43/0.9273 & 30.34/0.8422 & 29.10/0.8050   & 28.00/0.8491    \\
		& RKD                     & 34.43/0.9274 & 30.33/0.8423 & 29.09/0.8051 & 27.96/0.8493 \\
		& AT                      & 34.29/0.9262 & 30.26/0.8406 & 29.03/0.8035 & 27.76/0.8443 \\
		& FitNet                  & 33.35/0.9178 & 29.71/0.8323 & 28.62/0.7949 & 26.61/0.8167 \\
		& FAKD              & 34.39/0.9272 &
  30.34/0.8426 &
  29.10/0.8052 &
  28.07/0.8511 \\
		& CSD              & 34.45/0.9275 & 30.32/0.8430  & 29.11/0.8061 & 28.21/0.8549 \\
		& \eblue{MipKD }     & \eblue{34.59/0.9287}             &    \eblue{30.48/0.8447}          & \eblue{29.19/0.8070}             &     \eblue{28.41/0.8571}                       \\ \midrule
		\multirow{9}{*}{x4}    & Teacher                 & 32.65/0.9005 & 28.95/0.7903 & 27.81/0.744  & 26.87/0.8086 \\
		& Scratch           & 32.29/0.8965 & 28.68/0.7840  & 27.64/0.7380  & 26.21/0.7893 \\
		& KD         & 32.30/0.8965  & 28.70/0.7842  & 27.64/0.7382 & 26.21/0.7897 \\
		& RKD                     & 32.30/0.8965  & 28.69/0.7842 & 27.64/0.7383 & 26.20/0.7899  \\
		& AT                      & 32.22/0.8952 & 28.63/0.7825 & 27.59/0.7365 & 25.97/0.7825 \\
		& FitNet                  & 31.65/0.8873 & 28.33/0.7768 & 27.38/0.7309 & 25.40/0.7637  \\
		& FAKD              & 32.27/0.8960 &
  28.65/0.7836 &
  27.62/0.7379 &
  26.18/0.7895 \\
		& CSD              & 32.34/0.8974 & 28.72/0.7856 & 27.68/0.7396 & 26.34/0.7948 \\
		& \eblue{MipKD}                   & \eblue{32.46/0.8981} & \eblue{28.79/0.7863} & \eblue{27.71/0.7400}   & \eblue{26.45/0.7960}  \\ \bottomrule
	\end{tabular}\vspace{-0.5em}
\end{table}

\begin{table}[htb]
\caption{Quantitative comparison on RCAN~\cite{zhang2018image} architecture on the benchmark datasets. In these experiments, the RCAN student model of c64b6 is distilled by the teacher model of c64b20.}
\centering
\label{tab:mipkd-rcan}
	\begin{tabular}{@{}l@{\hspace{3\tabcolsep}}l@{\hspace{3\tabcolsep}}c@{\hspace{3\tabcolsep}}c@{\hspace{3\tabcolsep}}c@{\hspace{3\tabcolsep}}c@{}}
        \toprule
		\multirow{2}{*}{Scale} & \multirow{2}{*}{Method} & Set5         & Set14        & BSD100       & Urban100     \\ \cmidrule(l){3-6} 
		&                         & PSNR/SSIM    & PSNR/SSIM    & PSNR/SSIM    & PSNR/SSIM    \\ \midrule
		\multirow{8}{*}{x2}    & Teacher                 & 38.27/0.9614 & 34.13/0.9216 & 32.41/0.9027 & 33.34/0.9384 \\
		& Scratch                 & 38.13/0.9610  & 33.78/0.9194 & 32.26/0.9007 & 32.63/0.9327 \\
		& KD                      & 38.17/0.9611 & 33.83/0.9197 & 32.29/0.9010  & 32.67/0.9329 \\
		& RKD                     & 38.18/0.9612 & 33.78/0.9191 & 32.29/0.9011 & 32.70/0.9330   \\
		& AT                      & 38.13/0.9610  & 33.70/0.9187  & 32.25/0.9005 & 32.48/0.9313 \\
		& FitNet                  & 37.97/0.9602 & 33.57/0.9174 & 32.19/0.8999 & 32.06/0.9279 \\
		& FAKD                    & 38.17/0.9612 &
  33.83/0.9199 &
  32.29/0.9011 &
  32.65/0.9330  \\
		& \eblue{MiPKD}                   & \eblue{38.21/0.9613} & \eblue{33.92/0.9203} & \eblue{32.32/0.9015} & \eblue{32.83/0.9344} \\ \midrule
		\multirow{8}{*}{x3}    & Teacher                 & 34.74/0.9299 & 30.65/0.8482 & 29.32/0.8111 & 29.09/0.8702 \\
		& Scratch                 & 34.61/0.9288 & 30.45/0.8444 & 29.18/0.8074 & 28.59/0.8610  \\
		& KD                      & 34.61/0.9291 & 30.47/0.8447 & 29.21/0.8080  & 28.62/0.8612 \\
		& RKD                     & 34.67/0.9292 & 30.48/0.8451 & 29.21/0.8080  & 28.60/0.8610   \\
		& AT                      & 34.55/0.9287 & 30.43/0.8438 & 29.17/0.8070  & 28.43/0.8577 \\
		& FitNet                  & 34.21/0.9248 & 30.20/0.8399  & 29.05/0.8044 & 27.89/0.8472 \\
		& FAKD                    & 34.63/0.9290 &
  30.51/0.8453 &
  29.21/0.8079 &
  28.62/0.8612 \\
		& \eblue{MiPKD}                   & \eblue{34.72/0.9296} & \eblue{30.55/0.8458} & \eblue{29.25/0.8087} & \eblue{28.76/0.8640}  \\ \midrule
		\multirow{8}{*}{x4}    & Teacher                 & 32.63/0.9002 & 28.87/0.7889 & 27.77/0.7436 & 26.82/0.8087 \\
		& Scratch                 & 32.38/0.8971 & 28.69/0.7842 & 27.63/0.7379 & 26.36/0.7947 \\
		& KD                      & 32.45/0.8980 & 28.76/0.7860  & 27.67/0.7400   & 26.49/0.7982\\
		& RKD                     & 32.39/0.8974 & 28.74/0.7856 & 27.67/0.7399 & 26.47/0.7981 \\
		& AT                      & 32.31/0.8967 & 28.69/0.7839 & 27.64/0.7385 & 26.29/0.7927 \\
		& FitNet                  & 31.99/0.8899 & 28.50/0.7789  & 27.55/0.7353 & 25.90/0.7791  \\
		& FAKD                    & 32.46/0.8980 &
  28.77/0.7860 &
  27.68/0.7400 &
  26.50/0.7980 \\
		& \eblue{MiPKD}                   & \eblue{32.46/0.8982} & \eblue{28.77/0.7860}  & \eblue{27.69/0.7402} & \eblue{26.55/0.7998} \\ \bottomrule
	\end{tabular}\vspace{-1.5em}
\end{table}

We compare MiPKD with the baselines: train from scratch, Logits-KD~\cite{hinton2015distilling}, RKD~\cite{park2019relational}, AT~\cite{zagoruyko2016paying},  FitNet~\cite{Romero2014FitNetsHF}, FAKD~\cite{he2020fakd}, and CSD~\cite{wang2021towards}. Since the CSD is a self-distillation method in the channel-spliting manner, it's not applicable to the RCAN experiments of network depth distillation. 
The results for \texttimes4 EDSR trained with CSD are obtained by testing the provided checkpoint, and the \texttimes2 and \texttimes3 ones are reproduced by us since the checkpoints are unavailable.
To evaluate quality of SR model's output, we calculate the peak signal-to-noise ratio (PSNR) and the structural similarity index (SSIM) on the Y channel of the YCbCr color space. We use 800 images from DIV2K~\cite{timofte2017ntire} for training and evaluate SR models on four benchmark datasets: Set5~\cite{bevilacqua2012low}, Set14~\cite{zeyde2012single}, BSD100~\cite{martin2001database}, and Urban100~\cite{huang2015single}.%

\noindent\textbf{Training Details.} All models are trained using Adam~\cite{kingma2014adam} optimizer with $\beta_1 = 0.9$, $\beta_2 = 0.99$ and $\epsilon = 10^{-8}$, with a batch size of 16 and a total of $2.5\times 10^{5}$ updates. The initial learning rate is set to $10^{-4}$ and is decayed by a factor of 10 at every $10^{5}$ iteration. We set the loss weights $\lambda_1$ and $\lambda_2$ to 10 and 1, respectively. The proposed MiPKD is implemented by the BasicSR~\cite{basicsr} and PyTorch~\cite{paszke2019pytorch} framework and train them using 4 NVIDIA V100 GPUs. The LR images used for training and evaluation were obtained by down-sampling the HR images with the bicubic degradation method. During training, the input images are randomly cropped into $48\times 48$ patches and augmented with random horizontal/vertical flips and rotations.

\subsection{Results and Comparison}

\begin{table}[]
\caption{Quantitative comparison of distilling SwinIR~\cite{liang2021swinir} on the benchmark datasets.
}
\label{tab:swinir}
\centering
\begin{tabular}{@{}cccccccccc@{}}
\toprule
\multirow{2}{*}{Scale} & \multirow{2}{*}{Method} & \multicolumn{2}{c}{Set5} & \multicolumn{2}{c}{Set14} & \multicolumn{2}{c}{BSD100} & \multicolumn{2}{c}{Urban100} \\ \cmidrule(l){3-10} 
                       &                         & PSNR       & SSIM        & PSNR        & SSIM        & PSNR        & SSIM         & PSNR         & SSIM          \\ \midrule
\multirow{4}{*}{2}     & Teacher                 & 38.36      & 0.9620      & 34.14       & 0.9227      & 32.45       & 0.9030       & 33.40        & 0.9394        \\
                       & Scratch                    & 38.00      & 0.9607      & 33.56       & 0.9178      & 32.19       & 0.9000       & 32.05        & 0.9279        \\
                       & KD                      & 38.04      & 0.9608      & 33.61       & 0.9184      & 32.22       & 0.9003       & 32.09        & 0.9282        \\
                       & \eblue{MipKD}                   & \eblue{38.14}      & \eblue{0.9611}      & \eblue{33.76}       & \eblue{0.9194}      & \eblue{32.29}       & \eblue{0.9011}       & \eblue{32.46}        & \eblue{0.9313}        \\ \midrule
\multirow{4}{*}{3}     & Teacher    & 34.89    &   0.9312 & 30.77  &   0.8503 & 29.37  &   0.8124 & 29.29 & 0.8744  \\
                       & Scratch                    & 34.41      & 0.9273      & 30.43       & 0.8437      & 29.12       & 0.8062       & 28.20        & 0.8537        \\
                       & KD                      & 34.44      & 0.9275      & 30.45       & 0.8443      & 29.14       & 0.8066       & 28.23        & 0.8545        \\
                       & \eblue{MipKD}                   & \eblue{34.53}      & \eblue{0.9283}      & \eblue{30.52}       & \eblue{0.8456}      & \eblue{29.19}       & \eblue{0.8079}       & \eblue{28.47}        & \eblue{0.8591}        \\ \midrule
\multirow{4}{*}{4}     & Teacher                 & 32.72      & 0.9021      & 28.94       & 0.7914      & 27.83       & 0.7459       & 27.07        & 0.8164        \\
                       & Scratch                    & 32.31      & 0.8955      & 28.67       & 0.7833      & 27.61       & 0.7379       & 26.15        & 0.7884        \\
                       & KD                      & 32.27      & 0.8954      & 28.67       & 0.7833      & 27.62       & 0.7380       & 26.15        & 0.7887        \\
                       & \eblue{MipKD}                   & \eblue{32.39}      & \eblue{0.8971}      & \eblue{28.76}       & \eblue{0.7854}      & \eblue{27.68}       & \eblue{0.7403}       & \eblue{26.37}        & \eblue{0.7956}        \\ \bottomrule
\end{tabular}
\end{table}

\noindent\textbf{Comparison with Baseline Methods.}  
Quantitative results for training EDSR~\cite{lim2017enhanced}, RCAN~\cite{zhang2018image}, and SwinIR~\cite{liang2021swinir} of three SR scales 
are presented in~\cref{tab:mipkd-edsr}, \cref{tab:mipkd-rcan} and~\cref{tab:swinir}, from which we can draw the following conclusions: 

\textbf{(1)} Existing KD methods for SR have limited effects, some may even deteriorate the student model. The KD methods originally designed for high-level CV tasks (RKD, AT, FitNet), though applicable, hardly improve the SR models over training from scratch. For instance, AT and FitNet underperform the vanilla student models trained without KD among all settings. 

\textbf{(2)} The presented MiPKD outperforms existing KD methods baseline in width and depth compression, respectively. 
For example, MiPKD outperforms the vanilla student in the most challenging dataset Urban100 in EDSR$\times$2, $\times$3, and $\times$4 settings by
\textbf{0.56 dB}, \textbf{0.42 dB}, \textbf{0.24 dB} 
in terms of PSNR, respectively as~\cref{tab:mipkd-edsr} shown. %
For depth distillation, compared with trainning from scratch,
\textbf{0.2 dB}, \textbf{0.17 dB}, \textbf{0.19 dB} 
in terms of PSNR are improved, respectively, on Urban100 dataset in RCAN $\times$2, $\times$3, and $\times$4 settings as~\cref{tab:mipkd-rcan} shown.

\textbf{(3)} The MiPKD is applicable to Transformer network and able to boost the model's performance. Conventional feature-based KD methods are not directly applicable to the Transformer-type networks, so we compare MiPKD with training from scratch and the response-based KD~\cite{hinton2015distilling} in the experiments. The results in ~\cref{tab:swinir} indicate that the MiPKD could improve the transformer SR model by a large margin, further emphasizing its superior performance.

\begin{table}[]
\caption{Quantitative comparison on training \texttimes 4 EDSR~\cite{lim2017enhanced} models with higher compression rate on the benchmark datasets.In these experiments, the EDSR student model of c64b16 is distilled by the teacher model of c256b32.}
\label{tab:mipkd-edsr-high-cmpr-rate}
\centering
	\begin{tabular}{@{}l@{\hspace{3\tabcolsep}}c@{\hspace{3\tabcolsep}}c@{\hspace{3\tabcolsep}}c@{\hspace{3\tabcolsep}}c@{}}  
		\toprule
		\multirow{2}{*}{Method} & Set5         & Set14        & BSD100       & Urban100     \\ \cmidrule(l){2-5} 
		& PSNR/SSIM    & PSNR/SSIM    & PSNR/SSIM    & PSNR/SSIM    \\ \midrule
		Teacher                 & 32.65/0.9005 & 28.95/0.7903 & 27.81/0.7440  & 26.87/0.8086 \\
		NOKD                    & 32.01/0.8924 & 28.46/0.7782 & 27.47/0.7324 & 25.61/0.7704 \\
		KD                      & 31.99/0.8921 & 28.46/0.7784 & 27.47/0.7327 & 25.60/0.7700    \\
		FitNet                  & 31.92/0.8912 & 28.42/0.7776 & 27.44/0.7317 & 25.52/0.7672 \\
		FAKD                    & 31.65/0.8879 & 28.32/0.7760  & 27.37/0.7303 & 25.38/0.7629 \\
		FAKD+CSD                & 32.00/0.8930     & 28.47/0.7800   & 27.51/0.7340  & 25.79/0.7790  \\
		CSD+FAKD                & 31.86/0.8907 & 28.42/0.7786 & 27.46/0.7327 & 25.58/0.7709 \\
		\eblue{MiPKD+MiPKD}             & \eblue{32.17/0.8947}            & \eblue{28.57/0.7812}             & \eblue{27.57/0.7354}             & \eblue{25.89/0.7794}              \\ \bottomrule
	\end{tabular}
\end{table}

\noindent\textbf{Evaluate MiPKD on high compression rate setting.} \cref{tab:mipkd-edsr-high-cmpr-rate} shows the results of training \texttimes4 scale EDSR model compressed in both network width and depth dimensions. The number of parameters are compressed for about 28 times from teacher model.
Directly distilling the student model by the teacher model yields negative effects on its performance. Two-stage KD with an intermediate teaching-assistant (TA) model~\cite{mirzadeh2020improved} are necessary in such case. 
To make use of the CSD method, we compared different TA options, (1) Teacher->Student1->Student2: the whole pipeline is in a width-then-depth compression order. We adopt CSD to train the TA and FAKD for student. model (2) Depth-then-width: we first perform depth compression and then width compression. We adopt FAKD+CSD and MiPKD+MiPKD for the two-stage distillations. Using MiPKD to train both TA and student model yields the best results.

\begin{figure*}[htb]
		\centering
		\includegraphics[width=\linewidth]{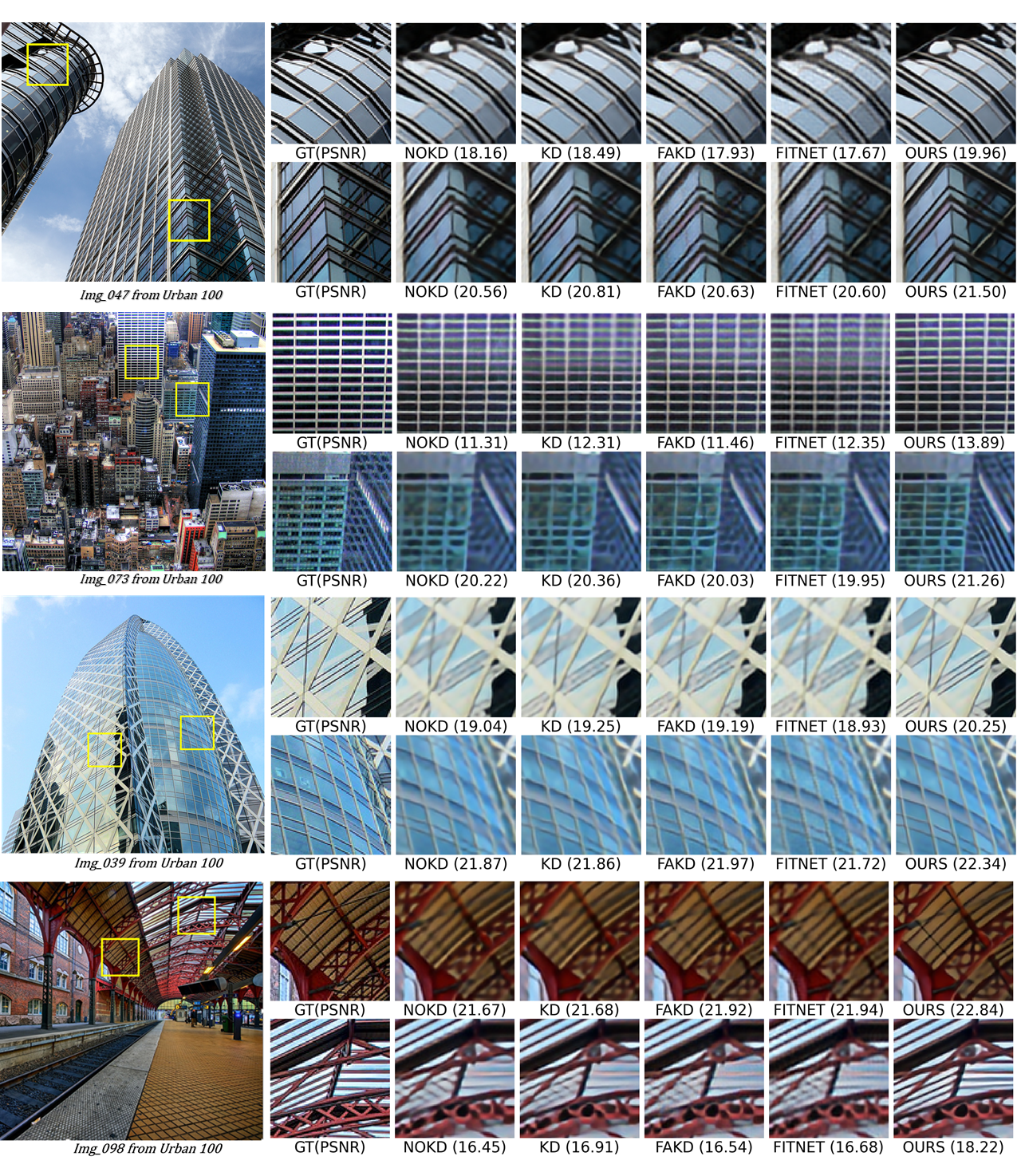}
		\vspace{-1em}
		\caption{Visual comparison ($\times$4) with existing SRKD methods from Urban100. the numbers in the bracket denote the PSNR of the presented patches.
		}
		\label{fig:Fig-2}
	\end{figure*}

\noindent\textbf{Visual Comparison}.~\cref{fig:Fig-2} compares the output of \texttimes4 RCAN models differently trained for four images from the Urban100 dataset. For instance, for $img\_047$, MiPKD can reconstruct much better fine details than all baseline work. FAKD are prone to artifacts in the left-bottom of the building and the vanilla student, Logits-KD, FAKD, and FitNet are over-blurred. In contrast, MiPKD alleviates the blurring artifacts and reconstructs much more structural details. Similar observations can be found in other cases, \textit{e.g.} the characters and anisotropic textures in $img\_073$. These visual comparisons are consistent with the quantitative results, demonstrating the superiority of MiPKD. More visual comparisons can be found in the supplementary materials.

\section{Ablation Study}\label{sec:ablation}

To demonstrate the effectiveness of the proposed MiPKD scheme, we conduct detailed ablation studies on \texttimes multiple scale RCAN and EDSR networks.

\noindent\textbf{Ablation on the feature and block prior mixers for MipKD.}
There are two fine-grained prior mixer modules in MiPKD, namely, the feature and block prior mixers. Their individual effects are ablated in~\cref{tab:ablation-mipkd-mixer}. The result shows that  employing the feature prior mixers leads to significant performance improvement and the block prior mixer based on it could further boost the student model.

\begin{table}[]
\caption{Ablation on the two modules of MiPKD. In these experiments, the RCAN student model of c32b5g5 is distilled by the teacher model of c32b6g10. Applying both of the mixers together shows the best student model performance.}\vspace{-0.5em}
\label{tab:ablation-mipkd-mixer}
\centering
\resizebox{0.7\columnwidth}{!}{%
\begin{tabular}{@{\hspace{2pt}}ccc@{\hspace{2pt}}}
\toprule
\multirow{2}{0.25\columnwidth}{\centering Feature Prior Mixer} & \multirow{2}{0.25\columnwidth}{\centering Block Prior Mixer} & Urban100 \\ \cmidrule(lr){3-3} 
&    & PSNR / SSIM \\ \midrule
\xmark                                    & \xmark                                  & 25.60 / 0.7700           \\
\cmark                                    & \xmark                                  & 25.63 /   0.7711 \\
\xmark                                    & \cmark                                  & 25.65 /   0.7717       \\
\cmark                                    & \cmark                                  & 25.69 /    0.7728       \\ \bottomrule
\end{tabular} \vspace{-1em}
}
\end{table}

\begin{table}[]
\caption{Ablation on the encoder type in MiPKD feature mixer module without block prior mixer module.}\vspace{-0.5em}
\centering
\label{tab:ablation-mipkd-encoder}
\resizebox{0.35\columnwidth}{!}{%
\begin{tabular}{@{\hspace{2pt}}lcc@{\hspace{2pt}}}
\toprule
\multirow{2}{*}{Encoder Type} & \multicolumn{2}{c}{Urban100} \\ \cmidrule(lr){2-3} 
                              & PSNR         & SSIM          \\ \midrule
No Encoder                    & 24.51        & 0.7149       \\
Shared Encoder                & 25.61        & 0.7704        \\
Separate Encoder              & 25.63        & 0.7711             \\ \bottomrule
\end{tabular}%
}
\end{table}

\noindent\textbf{Ablation on the MiPKD feature prior mixer module.} 
In the feature prior mixer module of MiPKD, the teacher and student models' feature maps are mapped to the latent space through corresponding encoders, thereby randomly mixed and stitched. 
We present an analysis of the impact of the encoders in~\cref{tab:ablation-mipkd-encoder} by comparing MiPKD with 1) removing the encoders, aligning and utilizing the teacher's feature map directly and 2) sharing a encoder among the teacher and student model. 
Removing the encoders would substantially deteriorate the student model's performance. Due to the different distribution of teacher and student models' feature maps, a shared encoder cannot effectively map them to the same latent space, leading to noisy mixtures.
Assigning the teacher and student models separate encoders yields the best results, indicating that mixing feature priors in the same latent space is necessary.

\cref{tab:ablation-mipkd-encoder2} compares the encoders and decoder of different architecture. The result shows that convolutional neural network type exhibits better performance and is more suitable in SR task.

Besides, the strategies of the mask generation are compared in~\cref{tab:ablation-mipkd-mask}. Compared with 1) masking according to  the Cosine or CKA similarity between teacher and student models' feature maps or 2) generating the complementary pairs of feature map by fixed grid pattern, the random 3D-mask exhibits the best performance and least calculation consumption. A more flexible, generalisable strategy is applied in the prior mixer module.

\begin{table}[]

\caption{Ablation analysis on the encoder/decoder network architecture settings. The RCAN student model of c64b5g10 is distilled by the teacher model of c64b20g10.}
 \centering
 \label{tab:ablation-mipkd-encoder2}
	\begin{tabular}{@{}lcccc@{}}
		\toprule
		\multirow{2}{*}{Encoder Type} & Set5         & Set14        & BSD100       & Urban100     \\ \cmidrule(l){2-5} 
		& PSNR/SSIM    & PSNR/SSIM    & PSNR/SSIM    & PSNR/SSIM    \\ \midrule
		MLP                           & 32.39/0.8976 & 28.72/0.7853 & 27.65/0.7391 & 26.42/0.7964 \\
		Conv                          & 32.46/0.8982 & 28.77/0.7860  & 27.69/0.7402 & 26.55/0.7998 \\ \bottomrule
	\end{tabular}\vspace{-1em}
\end{table}

\begin{table}[]
\caption{Ablation analysis on the masking strategy for feature prior mixture.}
\label{tab:ablation-masking-strategy}
 \centering
 \label{tab:ablation-mipkd-mask}
	\begin{tabular}{@{}lc@{}}
		\toprule
		\multirow{2}{*}{masking strategy} & Urban100     \\
		& PSNR/SSIM    \\ \midrule
		Cosine                       & 25.62/0.7711 \\
		Grid mask                    & 25.61/0.7669 \\
		CKA                           & 25.63/0.7713 \\
		Random                        & 25.69/0.7728 \\ \bottomrule
	\end{tabular}
\end{table}

\noindent\textbf{Ablation on the ``auto-encoder'' loss $L_k^{ae}$. } We compared the MiPKD with and without $L_k^{ae}$ in~\cref{tab:ablation-autoencoder-loss}. The results indicate that the auxiliary ``auto-encoder'' loss makes the mapping between original feature maps' space and the latent space more accurate, leading to better student model's performance. 

\begin{table}[]\vspace{-1.5em}
\caption{Ablation study on the auto-encoder loss.}
 \centering
 \label{tab:ablation-autoencoder-loss}
	\begin{tabular}{@{}ccccc@{}}
		\toprule
		\multirow{2}{*}{Auto-encoder Loss} & Set5         & Set14        & BSD100       & Urban100     \\ \cmidrule(l){2-5} 
		& PSNR/SSIM    & PSNR/SSIM    & PSNR/SSIM    & PSNR/SSIM    \\ \midrule
		\xmark & 32.4/0.8976  & 28.71/0.7851 & 27.66/0.7396 & 26.42/0.7971 \\
		\cmark & 32.46/0.8982 & 28.77/0.7860 & 27.69/0.7402 & 26.55/0.7998 \\ \bottomrule
	\end{tabular}\vspace{-1.5em}
\end{table}

\noindent
\textbf{Ablation on the Loss weights setting of feature and block mixers.}
The impact of various weights of feature mixers loss and block mixer loss is evaluated as Tab.\ref{tab:ablation-loss-weight} shown, where $\lambda_{rec}$, $\lambda_{kd}$, $\lambda_{feat}$, $\lambda_{block}$  represent the weights for reconstruction loss, logits-kd loss, feature prior mixer and block prior mixer respectively.
Considering the initial fluctuation caused by mixing the block from networks, $\lambda_{block}$ is applied as 0.1 presents the best student performance as the Tab.\ref{tab:ablation-autoencoder-loss} shown.
In addition, the reconstruction loss of auto-encoder in the feature prior mixer is introduced in the initial stage of training.
As the reconstruction ability of decoder improves, it's beneficial for the prior mixer to fuse dark knowledge and restore the enhanced feature map efficiently.

\begin{table}[]\vspace{-1.5em}
\caption{Ablation analysis on the weights of different losses}
 \centering
 \label{tab:ablation-loss-weight}
	\begin{tabular}{@{}cccccccc@{}}
		\toprule
		\multirow{2}{*}{$\lambda_{rec}$} & \multirow{2}{*}{$\lambda_{kd}$} & \multirow{2}{*}{$\lambda_{feat}$} & \multirow{2}{*}{$\lambda_{block}$} & Set5         & Set14        & BSD100       & Urban100     \\ \cmidrule(l){5-8} 
		&                                 &                                   &                                    & PSNR/SSIM    & PSNR/SSIM    & PSNR/SSIM    & PSNR/SSIM    \\ \midrule
		1                                & 1                               & 1                                 & 1                                  & 32.44/0.8972 & 28.75/0.7851 & 27.68/0.7399 & 26.51/0.7976 \\
		1                                & 1                               & 0.1                               & 1                                  & 32.34/0.8970 & 28.73/0.7849 & 27.67/0.7394 & 26.47/0.7960 \\
		1                                & 1                               & 0.1                               & 0.1                                & 32.42/0.8980 & 28.75/0.7857 & 27.68/0.7399 & 26.51/0.7988 \\
		1                                & 1                               & 1                                 & 0.1                                & 32.46/0.8982 & 28.77/0.7860 & 27.69/0.7402 & 26.55/0.7998 \\ \bottomrule
	\end{tabular}\vspace{-1.5em}
\end{table}

\section{Conclusion}
In this paper, we present MiPKD, a simple yet significant KD framework for SR with feature and block mixers. 
By incorporating the prior knowledge from teacher model to the student, the capacity disparity between them are reduced, and the feature alignment is achieved effectively.  
We compare the proposed MiPKD with various kinds of distillation methods and demonstrate the superiority of our solutions. 
For instance, it improves the EDSR network of three SR scale on Urban100 testing set in the range of  \textbf{0.24 dB} $\sim$ \textbf{0.56 dB} in terms of PSNR. 
The presented MiPKD outperforms existing KD methods baseline in width, depth compression and compounded compression in SR tasks, which is flexible to a wide array of teacher-student kd architectures.

\FloatBarrier
\newpage
\bibliographystyle{splncs04}
\bibliography{main}
\end{document}